%% file: report.tex
\documentclass[conference, letterpaper]{IEEEtran}

\usepackage[utf8]{inputenc}
\usepackage[cmex10]{amsmath}
\usepackage{amssymb, mathtools}
\usepackage[pdftex]{graphicx}
\usepackage[caption=false,font=footnotesize]{subfig}
\usepackage[most]{tcolorbox}
\usepackage{hyperref}
\usepackage[ruled,vlined,linesnumbered]{algorithm2e}
\usepackage{textcomp}
\usepackage{fancyhdr}
\usepackage{listings}
\usepackage{tikz}
\usepackage{xurl} 
\usepackage{url}
\usepackage{amsfonts}
\usepackage{tabularx}
\usepackage{booktabs}
\usepackage{multirow}
\usepackage{forest}
\usepackage{placeins}
\usepackage{enumitem}
\usepackage{pgfplots}
\usepackage{xcolor}
\usepackage{pgf-pie}
\usepackage{adjustbox}
\usepackage{titlesec}
\usepackage{mathptmx}
\usepackage{float}
\usepackage{threeparttable}
\usepackage{pdflscape}

\definecolor{myblockcolor}{RGB}{110,0,25}
\definecolor{hkustblue}{RGB}{120,5,30}
\definecolor{hkustyellow}{RGB}{110,0,25}
\definecolor{colorOneLight}{RGB}{140,70,90}
\titlespacing*{\section}{0pt}{0.8\baselineskip}{0.4\baselineskip}
\makeatletter
\def\ALG@name{Algorithm}
\makeatother
\pgfplotsset{compat=1.18}
\usetikzlibrary{arrows.meta, positioning, calc, fit, shapes.geometric}
\newcommand\tab[1][0.5cm]{\hspace*{#1}}

\newcommand{\PlaysRole}{\mathsf{PlaysRole}}

\newcommand{\DifferentFrom}{\mathsf{DifferentFrom}}

\newcommand{\cls}[1]{\ensuremath{\mathsf{#1}}}
\newcommand{\role}[1]{\ensuremath{\mathsf{#1}}}
\hypersetup{
    colorlinks=true,
    linkcolor=red,
    citecolor=blue,
    urlcolor=black,
    filecolor=black,
    linkbordercolor={0 0 0},
    pdfborder={0 0 0}
}
\allowdisplaybreaks
\PassOptionsToPackage{hyphens}{url}      
\renewcommand{\thispagestyle}[2]{}
\fancypagestyle{plain}{
    \fancyhead{}
    \fancyhead[C]{first page center header}
    \fancyfoot{}
    \fancyfoot[C]{first page center footer}
}
\makeatletter
\def\ALG@name{Algorithm}
\makeatother
\title{Ontology for Policing: Conceptual Knowledge Learning for Semantic Understanding and Reasoning in Law Enforcement Reports}
\author{
    \IEEEauthorblockA{
    \begin{tabular}{cc}
    \begin{tabular}{c}
    Anita Srbinovska\\
    Office of Business Intelligence\\
    Rochester Police Department\\
    Rochester, NY, USA\\
    \href{mailto:Anita.Srbinovska@CityofRochester.gov}{Anita.Srbinovska@CityofRochester.gov}
    \end{tabular}
    &
    \begin{tabular}{c}
    Jansen Orfan\\
    Department of Computer Science\\
    Rochester Institute of Technology\\
    Rochester, NY, USA\\
    \href{mailto:jrovcs@rit.edu}{jrovcs@rit.edu}
    \end{tabular}
    \\[0.10cm]
    \multicolumn{2}{c}{\vspace{0.35cm}}\\
    \begin{tabular}{c}
    Adrian Martin\\
    Office of Business Intelligence\\
    Rochester Police Department\\
    Rochester, NY, USA\\
    \href{mailto:Adrian.Martin@CityofRochester.gov}{Adrian.Martin@CityofRochester.gov}
    \end{tabular}
    &
    \begin{tabular}{c}
    Ernest Fokou\'e\\
    School of Mathematics and Statistics\\
    Rochester Institute of Technology\\
    Rochester, NY, USA\\
    \href{mailto:epfeqa@rit.edu}{epfeqa@rit.edu}
    \end{tabular}
    \end{tabular}
    }
}

\begin{document}

\maketitle

\input{abstract}

\input{introduction}

\input{background}

\input{data}

\input{methods}

\input{evaluation}

\input{discussion}

\input{conclusion}

\input{acknowledgments}

\bibliographystyle{IEEEtran}
\bibliography{references}

\clearpage

\input{appendix}

\end{document}

%% file: abstract.tex
\begin{abstract}
    Law enforcement reports contain structured fields and written narratives. However, many incident facts that are needed for review, police training, and investigations are in natural language and require manual reading. We propose a framework using symbolic methods for converting narratives into evidence--linked facts. Our objective is to measure the value of narratives to recover incident details only from the unstructured text and build temporal graphs with time cues and domain axioms. We achieve this by redacting personal identifiers, semantic parsing, predicate mapping to ontology, and reasoning. We evaluate the symbolic approach on $\textbf{450}$ property crime reports and a short human review. Of the extracted events from the system, $\textbf{54.1\%}$ had a confidence score $\ge$ 0.80 and $\textbf{93.7\%}$ were mapped through the PropBank$\rightarrow$VerbNet$\rightarrow$WordNet semantic path. $\textbf{100\%}$ agreement was reached on incident initiation, stolen items, and temporal cues and lower agreement for forced entry interpretation.
\end{abstract}


%% file: introduction.tex
\section{Introduction}\label{sec:Intro}

\noindent Law enforcement agencies rely on incident reports to document events, investigations and keep track of operations which have \textit{structured fields} such as coded categories, checkboxes, administrative data (case number, offense title, and statute) and \textit{natural language} written by police officers. The metadata is machine--readable and supports counting, filtering and record management. Agencies use law enforcement records management system data entry tools to input structured data into incident reports, so a clear narrative is expected from law enforcement personnel that organizes incident facts, which are typically in chronological order~\cite{MPD_ReportWriting}. However, many important details of an incident are only in the narrative. These include what happened and in what order, who was involved and how participants such as \textit{officer}, \textit{victim} and \textit{suspect} are described. \\
\tab Narrative details can also vary across reports. One report may begin with a $911$ call, while another may begin with an officer on patrol, and details such as vehicles, stolen items, or where and how someone entered a property may appear only in the free text. These details are harder to access systematically which creates a major bottleneck when investigators or analysts need to review large numbers of reports. In some \textit{public--release} settings these narratives are redacted which also adds to the challenge of recovering important information as well as keeping traceability to the source text. \\
\tab In this work we focus only on extracting event details, participants, and temporal ordering from police narratives that are often missing from structured data. We treat these as \textit{evidence--linked facts}, meaning facts that are grounded in the narrative and can be traced back to the report text. Extracting this information can make narratives more useful for systematic review, analysis and investigation which leads to the following research questions:
\begin{enumerate}[label={}, noitemsep]
    \item \textbf{RQ$\textbf{1.}$} Can redacted police narratives be converted into evidence--linked facts that can be traced back to the original sentences with symbolic natural language understanding (NLU) techniques?
    
    \item \textbf{RQ$\textbf{2.}$} Do redacted police narratives provide enough evidence for event and temporal information that may not be captured in the structured metadata?
\end{enumerate}
\begin{figure*}[h]
    \centering
    \setlength{\fboxsep}{0pt}
    \resizebox{0.98\textwidth}{!}{
    \begin{tikzpicture}[
            font=\normalsize,
            box/.style={
                draw=black!25, rounded corners=4mm, fill=white,
                minimum width=6.2cm, minimum height=1.55cm, align=center
            },
            circ/.style={
                draw=black!25, circle, fill=white,
                minimum size=2.7cm, align=center
            },
            arrow/.style={-{Latex[length=4mm]}, line width=1.05pt, draw=black!60}
        ]
    
        \node[box] (s1) {\textbf{1.\ Privacy and Extraction} \\ \large Narrative $\rightarrow$ Redacted Text + Entities};
        \node[box, right=1.25cm of s1] (s2) {\textbf{2.\ Semantic Layer} \\ \large Structured Event Semantics (AMR)};
        \node[box, right=1.25cm of s2] (s3) {\textbf{3.\ Analytical Outputs} \\ \large Queries + Auditable Inference};
    
        \node[circ, below=1.25cm of s2] (hub) {\textbf{Ontology} \\ \textbf{(OWL/DL)}};
    
        \draw[arrow] (s1) -- (s2);
        \draw[arrow] (s2) -- (s3);
    
        \draw[arrow] (s1.south east) -- (hub.north west);
        \draw[arrow] (s2.south) -- (hub.north);
        \draw[arrow] (hub.north east) -- (s3.south west);
    \end{tikzpicture}
    }
    
    \vspace{-0.4em}
    \caption{Narrative redaction, extraction, AMR semantic normalization, and ontology outputs.}
    \label{fig:pipeline_overview}
    \vspace{-0.6em}
\end{figure*}
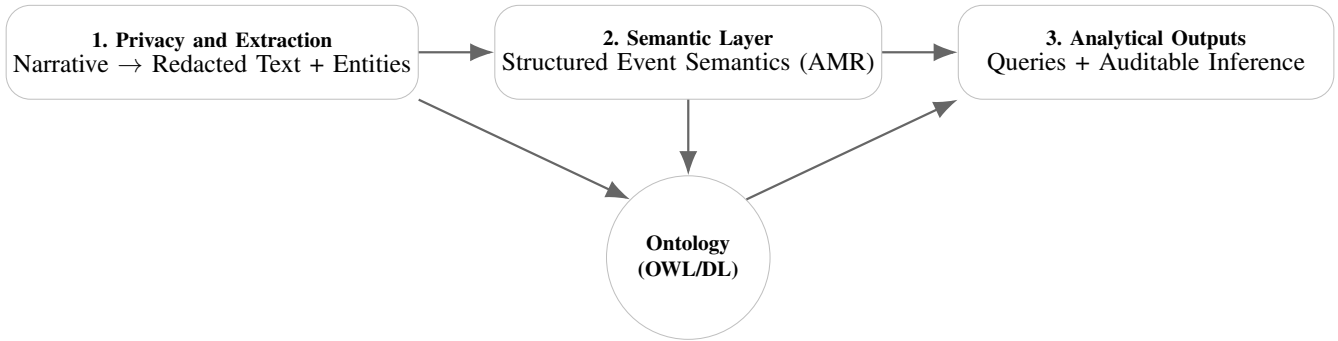
\noindent \tab For large--scale use, methods for analyzing police narratives must keep \textit{privacy}, handle \textit{linguistic variation} and maintain \textit{traceability} to the narratives. Figure~\ref{fig:pipeline_overview} illustrates the proposed symbolic approach. First, the narratives are redacted, and entities/events are extracted. Next the text is converted to semantic representation that keeps all events and participants. And finally the extracted meaning is mapped to ontology for reasoning. With this, we address RQ$1$, while RQ$2$ is examined with a short human review. \\
\noindent \tab The remainder of this paper is organized as follows. Section~\ref{related_work} reviews related work on police narratives and redaction, as well as on semantic parsing and linguistic knowledge bases for NLU. Section~\ref{data} describes the text corpus. Section~\ref{method} outlines our methods. Section~\ref{eval} shows the results. Section~\ref{limits} discusses limitations, usage, next steps and Section~\ref{conclude} concludes.

%% file: background.tex
\section{Related Work}\label{related_work}

\noindent \textbf{Symbolic NLU.} This work uses symbolic NLU since our goal is to extract facts from police narratives that can be typed, checked and audited. Prior work has shown that robust NLU helps move beyond shallow string matching toward text representations that support reasoning~\cite{AllenNLU}. In our approach, we treat the task as knowledge extraction (KE) from redacted police narratives, following the NLU tradition of representing sentence meaning in a form that supports semantic interpretation before reasoning~\cite{AllenNLU2}. \\
\tab Some information in police narratives is not stated directly and needs background knowledge for interpretation. Prior work has identified the \textit{knowledge bottleneck} as one of the main issues in symbolic NLU~\cite{PhD} and has shown that definitions can support the recovery of implicit commonsense relations~\cite{Commonsense_Knowledge}. This is relevant to our approach because event interpretation often depends on conceptual relations that give a deeper understanding. \\ 
\noindent \textbf{Lexical resources.} For many natural language processing applications, semantically meaningful sentence structures are best represented in the form of predicate--argument structure, i.e., \textit{``who did what to whom''}, and our extraction pipeline depends on identifying such information from police narratives. Much progress has been made in creating resources for consistent role annotation, especially in PropBank (PB) for predicate senses (verbs) and roles (arguments), which are often annotated with labels such as \texttt{:ARG0} or \texttt{:ARG1}~\cite{PalmerGildeaKingsbury2005PropBank}. These resources support our semantic parsing that recovers event structure. \\
\tab In our setting, we use lexical resources for event interpretation and argument typing. In particular, VerbNet (VN) organizes verbs into semantically related classes~\cite{VerbNetKipper2006} and WordNet (WN) is organized by synonym sets and hypernym relations~\cite{WordNetMiller1995}. We use VerbNet through SemLink~\cite{semlink} to connect predicate senses to semantic classes and WordNet that supports \textit{``is--a''} checks over extracted arguments and helps determine types such as vehicles, structures or structure parts. \\
\noindent \textbf{Semantic layer.} As shown in Figure~\ref{fig:pipeline_overview}, our workflow parses narratives into Abstract Meaning Representation (AMR) as an intermediate semantic layer. AMR represents sentence meaning as a graph of predicates and their participants, and prior work has studied sentence meaning and the challenges of accurate parsing~\cite{Survey,AMR_Parsing}. In our system, AMR provides the semantic input for the ontology. \\
\tab While FrameNet provides a way to represent event structure and frame semantics~\cite{FrameNet}, we use AMR graphs because they give explicit graph structure and PropBank sense labels that integrate directly with our mapping rules. In general, an event is an action or occurrence and a frame is a record that organizes the details of that event into groups or slots. For example an \cls{Entry} frame may have entry point, method, structure and tool slots and a \cls{Theft} frame may have stolen items and value mentioned. So this makes AMR a better fit for our pipeline where sentence meaning must be mapped into entities, events and roles. \\
\noindent \textbf{Ontologies.} In the Semantic Web community, ontologies give a formal schema for encoding constraints that support validation and reasoning~\cite{Event_Ontologies}, so after semantic parsing, we map extracted entities and events into an ontology. Prior work on ontologies in the policing domain has further shown that explicit schemas and logical constraints can be developed for property crime concepts~\cite{Property_Crimes}, which motivates our use of such representations for auditable police narrative analysis. \\
\noindent \textbf{Temporal event ordering.} Part of this work involves ordering events in time so investigators have a clearer picture of the timeline of events, e.g., the sentence \textit{``a suspect broke a window before entering a home''} shows a framework of temporal relationships~\cite{AllenFerguson_EventsActions}. Event representations have also been implemented in cognitive systems where some future states are dependent on past events~\cite{Cognitive_Agent}. While we do not implement the full framework, we do borrow some ideas to help extract and verify temporal relationships between events, as reflected in Section~\ref{sec:method:temporal}. \\
\noindent \textbf{Police narratives and redaction.} Police reports have fixed fields with written narratives by law enforcement officers. We focus only on narratives because they often contain important details that may not be fully captured in structured fields. However, operating at scale with narratives is difficult because incident reports can be incomplete or inconsistent~\cite{ProblemsPoliceReports}. In earlier studies, it was argued that textual analysis alone would not be sufficient for real--world challenges~\cite{Entities}. Recent work in this area has applied text mining and machine learning to other unstructured crime incident narratives (e.g., court documents) to address classification tasks (e.g., offense type). While we do not use machine learning in our approach, these studies show the value of free text in incident reports, while also suggesting that deep patterns cannot be captured from surface structure alone~\cite{ML}. \\
\tab Personally Identifiable Information (PII) may be in police narratives and since policing is a sensitive domain a typical preprocessing step is redaction before semantic parsing, ontology population and auditing~\cite{Redacting}. Hence, redaction is a necessary step in our workflow. \\
\noindent \textbf{OpenBWC.} This work extends \textit{OpenBWC}\footnote{OpenBWC is a research open--source initiative for ethical AI and statistical analysis of body--worn camera (BWC) footage: \texttt{\url{https://openbwc.org/}}.}, which is a collaboration between Rochester Institute of Technology, the Rochester Police Department (RPD), and criminologists at the University at Albany, by adding symbolic NLU components for internal use with the RPD through an ontology pipeline for redacted police narratives~\cite{srbinovska2025openbwc}.

%% file: data.tex
\section{Dataset and Preprocessing}\label{data}

\noindent This work uses a text corpus of $450$ RPD incident reports from five offense categories, \texttt{Burglary}, \texttt{Larceny}, \texttt{Motor Vehicle Theft}, \texttt{Robbery} and \texttt{Stolen Property}, between $2014$ and $2025$. The reports were provided in unredacted form. We only work with property crimes because their narratives often describe how the incident began and include useful details about individuals, vehicles, and objects. 
\begin{table}[h]
    \centering
    \caption{Dataset composition and narrative statistics.}
    \label{tab:data_composition}
    \small
    \begin{adjustbox}{max width=\columnwidth}
        \begin{tabular}{lrrrrr}
            \toprule
            \textbf{Category} & \textbf{\# Reports} & \textbf{Total Words} & \textbf{Avg. Words} & \textbf{Min} & \textbf{Max} \\
            \midrule
            Burglary & $138$ & $40{,}969$ & $297$ & $83$ & $846$ \\
            Larceny & $236$ & $40{,}078$ & $170$ & $40$ & $713$ \\
            Motor Vehicle Theft & $31$ & $8{,}182$ & $264$ & $65$ & $868$ \\
            Robbery & $38$ & $16{,}239$ & $427$ & $184$ & $694$ \\
            Stolen Property & $7$ & $1{,}137$ & $162$ & $89$ & $298$ \\
            \midrule
            \textbf{Total} & $450$ & $106{,}605$ & $237$ & $40$ & $868$ \\
            \bottomrule
        \end{tabular}
    \end{adjustbox}
\end{table}

\noindent \tab Table~\ref{tab:data_composition} shows a breakdown of the corpus by category with word count statistics. The narrative length is different from $40$ to $868$ words. This is important because more dense narratives have more entity and event descriptions. \\
\noindent \tab Before analysis, we perform the following steps:
\begin{enumerate}[noitemsep]
    \item \textbf{Extraction:} the source PDFs are processed to extract the \texttt{NARRATIVE} section, which is then transformed to plain text. We perform Optical Character Recognition (OCR), which is rendered at $300$ DPI and passed through Tesseract~\cite{Tesseract}, where heading, footer, and artifacts are removed. The extracted text in all uppercase is then converted to sentence case.
    
    \item \textbf{OCR:} the system corrects errors; for example we replace \texttt{|} with \texttt{I}.
    
    \item \textbf{Redaction:} we use spaCy's named entity recognition\footnote{spaCy API: \texttt{\url{https://spacy.io/api}}}, regular expressions and metadata to redact PII for example, names, addresses, dates of birth and vehicle details. We keep shorthand notations (\texttt{V} (Victim), \texttt{S} (Suspect), \texttt{W} (Witness)), use the same placeholders such as \texttt{[PERSON\_$\texttt{1}$]} for referencing the same entities and we also consider first--person mentions for reporting officers. This outputs redacted narrative files and audit files in JSON format, which record the locations of all placeholders.
\end{enumerate}
\begin{table}[h]
    \centering
    \caption{Redaction summary from audit logs across the corpus.}
    \label{tab:redaction_summary}
    \small
    \setlength{\tabcolsep}{5pt}
    \renewcommand{\arraystretch}{0.95}
    \begin{threeparttable}
    \begin{tabular}{lrr}
        \toprule
        \textbf{Redaction type} & \textbf{Total placeholders} & \textbf{Avg./report} \\
        \midrule
        PERSON & $7{,}295$ & $16.21$ \\
        DATE & $2{,}093$ & $4.65$ \\
        ORG & $1{,}030$ & $2.29$ \\
        ADDRESS & $668$ & $1.48$ \\
        {GPE\tnote{\textcolor{red}1}} & $228$ & $0.51$ \\
        PLATE & $56$ & $0.12$ \\
        LOC & $28$ & $0.06$ \\
        PHONE & $26$ & $0.06$ \\
        \midrule
        \textbf{Total} & $11{,}424$ & $25.39$ \\
        \bottomrule
    \end{tabular}

    \begin{tablenotes}
        \footnotesize
        \item [1]GPE = Geopolitical entity in named entity recognition. It refers to geographical locations that also have a governing body, such as countries, cities, states, provinces, and municipalities.
    \end{tablenotes}
    \end{threeparttable}
\end{table}
\tab In Table~\ref{tab:redaction_summary} we report the total number of placeholders for the redacted entities and the average number of placeholders per report. \\ \\
\noindent \textbf{Unit of analysis.} The incident reports are analyzed as individual documents. Due to the sensitive data in the narratives, the corpus is not publicly available. All incident reports are maintained and processed in a secure and controlled research computing environment for batch execution and large--scale analysis~\cite{RC}.

%% file: methods.tex
\section{Methodology}\label{method}

\noindent This paper proposes a framework for converting police narratives into evidence--linked facts through a symbolic pipeline. \\
\tab Algorithm~\ref{alg:symbolic_nlu_krr} summarizes the symbolic extraction pipeline. \\
\tab The algorithm begins by inducing an ontology $\mathcal{T}$ and a set of logic templates from the semantic descriptions $\mathcal{L}$ from the ontology $\mathcal{T}_0$ and a corpus $D_{\text{def}}$ (Line $1$). For example, a definition of theft can provide a template stating that a taking event with an agent, an item, and lack of permission supports a theft interpretation. \\
\tab For all reports $d \in \mathcal{D}$ the narrative $n$ and its metadata $m$ are extracted (Lines $3-4$). For example, the narrative may have \textit{``John Doe broke the window and took a wallet,''} and the metadata may include the case number, offense type and date. Then the narrative $n$ is redacted using the set of metadata rules $\mathcal{R}$ which results in redacted narrative $n'$ and redaction log $\ell_d$ (Lines $5-6$). Here \textit{``John Doe''} is replaced with \texttt{[PERSON\_1]}. \\
\tab Next, each sentence is parsed with the semantic parser $\mathcal{P}$ which gives a set of AMR graphs $G_d$ (Line $7$). For example the sentence ``\texttt{[PERSON\_1]} \textit{broke the window and took a wallet''} may produce predicate nodes such as \textit{break-$01$} and \textit{take-$01$} together with their arguments. The resulting AMR graphs are then transformed into extracted facts mapped to classes and roles $\mathcal{A}_N(d)$ (Line $8$), and separately the metadata is transformed into ontology facts $\mathcal{A}_M(d)$ (Line $9$) so it can give facts such as \cls{ForcedEntryEvent}, \cls{TheftEvent} or participant links for suspects. \\
\tab The narrative facts with the metadata--derived facts are merged into a case--level fact set $\mathcal{A}$ (Line $10$). A reasoning step $\mathcal{E}$ then checks the consistency of the fact set $\mathcal{A}$ with the applied ontology (Line $11$). For example, if a theft event is extracted, the reasoner can check whether a stolen item is present and whether participants are typed consistently. Inconsistent parts of the ontology are updated using a resolution step (Line $12$). \\
\tab For a single report, the algorithm outputs the redacted narrative, extracted events and participants, the evidence--linked facts, ontology and the validation output for audit (Line $13$).
\begin{algorithm}[h]
    \caption{\textbf{Symbolic KE}}
    \label{alg:symbolic_nlu_krr}
    \small
    \DontPrintSemicolon
    \SetKwInOut{Input}{Input}
    \SetKwInOut{Output}{Output}

    \Input{
        Reports $\mathcal{D}$; metadata $\mathcal{M}$; definition corpus $D_{\text{def}}$; linguistic KBs $\mathcal{K}_{\text{lex}}$; ontology $\mathcal{T}_0$; redaction rules $\mathcal{R}$; semantic parser $\mathcal{P}$; reasoner $\mathcal{E}$.
    }

    \Output{
        Redacted narratives $\mathcal{N}'$; narrative facts $\mathcal{A}_N$; metadata facts $\mathcal{A}_M$; rules $\mathcal{L}$; ontology $\mathcal{T}$; validation $\mathcal{V}$.
    }

    \tcp*[l]{Encode domain knowledge}
    $(\mathcal{T}, \mathcal{L}) \leftarrow \mathrm{\textbf{InduceLogic}}(D_{\text{def}}, \mathcal{T}_0)$\;

    \ForEach{$d \in \mathcal{D}$}{
        $n \leftarrow \mathrm{\textbf{ExtractNarrative}}(d)$\;
        $m \leftarrow \mathrm{\textbf{Metadata}}(d,\mathcal{M})$\;

        $(n', \ell_d) \leftarrow \mathrm{\textbf{Redact}}(n, m; \mathcal{R})$\;
        add $n'$ to $\mathcal{N}'$; store $\ell_d$\;

        $G_d \leftarrow \{ \mathcal{P}(s) \mid s \in \mathrm{\textbf{SentSplit}}(n') \}$\;
        $\mathcal{A}_N(d) \leftarrow \mathrm{\textbf{ExtractFacts}}(G_d; \mathcal{T}, \mathcal{L})$\;

        $\mathcal{A}_M(d) \leftarrow \mathrm{\textbf{MapMetadata}}(m, \mathcal{T})$\;
        $\mathcal{A} \leftarrow \mathcal{A}_N(d) \cup \mathcal{A}_M(d)$\;

        $\mathcal{V}(d) \leftarrow \mathcal{E}(\mathcal{T} \cup \mathcal{A})$\;

        $(\mathcal{T}, \mathcal{L}) \leftarrow \mathrm{\textbf{ResolveInconsistencies}}(\mathcal{T}, \mathcal{L}; \mathcal{V}(d), D_{\text{def}})$\;
    }

    \textbf{return} $\mathcal{N}', \mathcal{A}_N, \mathcal{A}_M, \mathcal{L}, \mathcal{T}, \mathcal{V}$\;
\end{algorithm}

\subsection{Knowledge Sources}

\noindent Police narratives contain semantics useful for agents, but they are often written with shorthand notation and varied sentence structures, so we employ a set of knowledge sources. \\
\tab We treat the extracted OWL/RDF\footnote{OWL: Web Ontology Language; RDF: Resource Description Framework} assertions as a \textit{per--report case knowledge base}. For each report $d$, the system constructs a case--level set of factual assertions containing event and participant instances linked to sentence evidence and to their extraction source. \\
\tab In our approach we have incorporated three different types of knowledge sources. Firstly, a participation--centric ontology $\mathcal{T}_0$ is used for structuring events that are linked to entities through participation roles, as described in Section~\ref{sec:method:owl}. Then, a definition (formal description of a concept) $D_{\text{def}}$ is used for inducting logic templates based on regularities that constrain the participation of typical individuals in events and relations between events. Finally, linguistic knowledge bases $\mathcal{K}_{\text{lex}}$ such as PropBank, VerbNet/SemLink and WordNet give lexical semantics to predicate normalization (mapping different verbs to event types) and argument typing. These lexical resources were accessed through the Natural Language Toolkit (NLTK) $3.9.2$~\cite{NLTKBook}, including PropBank, VerbNet, and WordNet $3.0$. \\ \\
\noindent \textbf{Semantic interpretation.} Semantic templates give some of this guidance for predicate normalization. We can use lexical semantic resources to provide information about events, roles, and additional constraints that a predicate is likely to express. \\
\tab Semantics can be determined by the main \textit{verb} which usually identifies the event and also from \textit{nouns} and \textit{arguments} that determine the participants and some other target objects. For example a \textit{take} event involving an agent, a transferred item and an impacted owner without authorization is the semantic meaning of theft predicate. In \textit{``The suspect stole a wallet from the victim’s vehicle,''} \textit{stole} is a theft event, \textit{wallet} is a stolen item, \textit{suspect} is the main agent, \textit{victim} is the affected entity and \textit{vehicle} is the source context:
\[
    \begin{aligned}
        & \textit{steal\_v1}(e) \wedge \textit{agent}(e, x) \wedge \textit{theme}(e, y) \wedge \textit{owner}(y, z) \\
        & \qquad \rightarrow \textit{take\_v1}(e) \wedge \textit{agent}(e, x) \wedge \textit{theme}(e, y) \\
        & \qquad\qquad \wedge \neg \textit{permission}(z, x, e).
    \end{aligned}
\]

\subsection{Semantic Parsing}\label{sec:method:semantic}

\noindent We use the AMRlib sentence--to--graph parser~\cite{Zhang2019AMRStoG}, which is a BART--large encoder--decoder model for AMR graphs with PropBank sense labels (e.g., \textit{steal-$01$}). The released checkpoint reports a SMATCH score of $83.7\%$. The AMR outputs are in PENMAN notation~\cite{Penman}, which is a text format for writing AMR graphs, so ontology population is reproducible. \begin{figure}[h]
    \centering
    \includegraphics[width=0.8\linewidth]{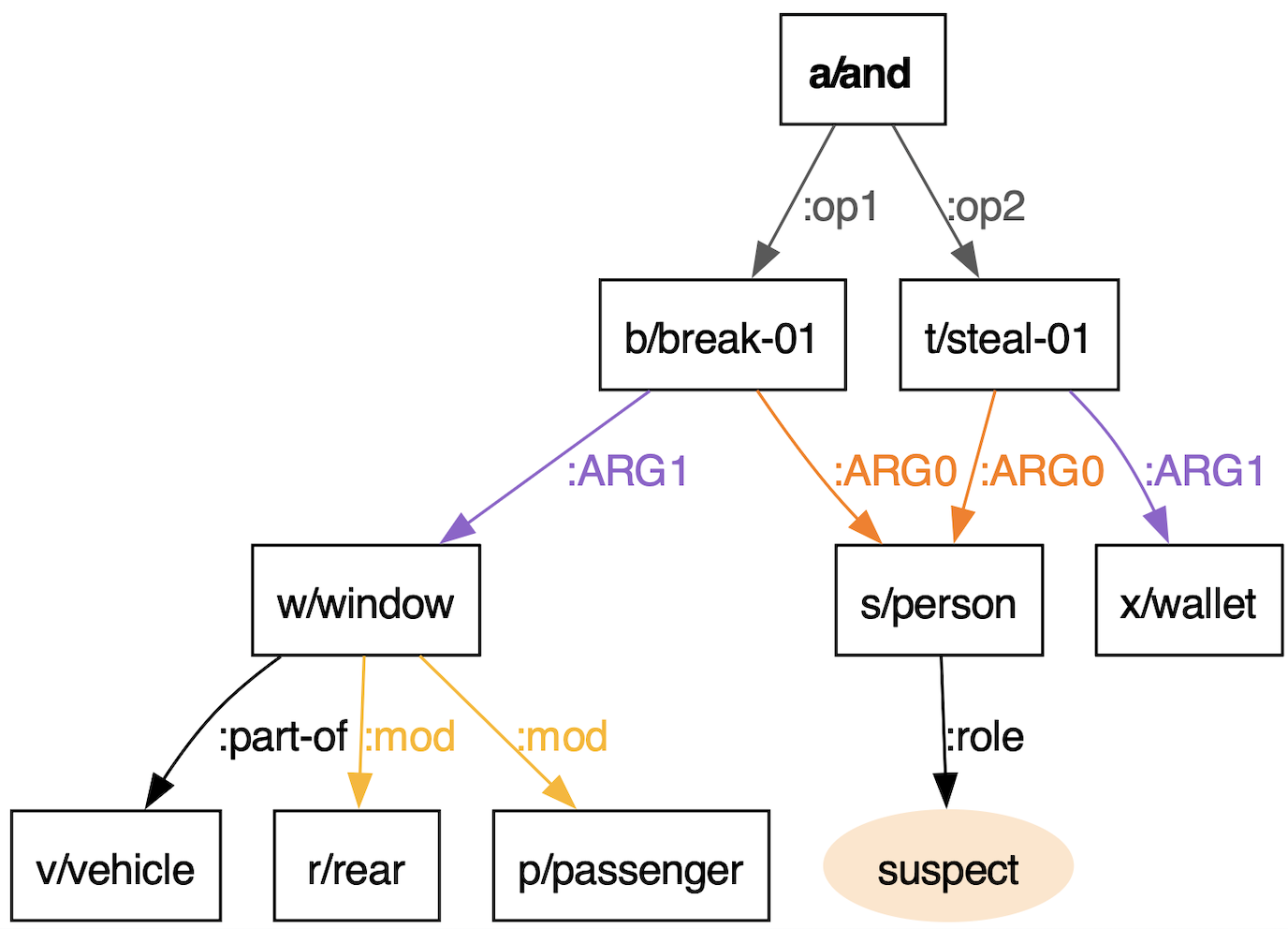}
    \caption{AMR graph for the sentence ``The suspect broke the rear passenger window of the vehicle and stole a wallet,'' showing two events (\textit{break-$01$} and \textit{steal-$01$}) with a shared agent (suspect).}
    \label{fig:placeholder}
\end{figure}

\noindent \tab Figure~\ref{fig:placeholder} shows how an AMR graph represents the meaning of a sentence. Each box corresponds to a concept node, where predicates such as \textit{break-$01$} and \textit{steal-$01$} represent events labeled with PropBank senses (``$01$'' indicating a specific sense of the verb). The edges between nodes are the semantic roles. For example, \texttt{:ARG0} usually denotes the agent (the doer of the action) and in this case is the \textit{person} (suspect), while \texttt{:ARG1} is the patient or object affected by the action, such as the \textit{window} in the breaking event and the \textit{wallet} in the stealing event. The other nodes represent the entities and their relationships. For example, \textit{window} is linked to \textit{vehicle} with a \textit{:part-of} relation and modifiers such as \textit{rear} and \textit{passenger} describe the object. The \textit{person} node connected to both predicates means that the same agent participates in both events. \\
\noindent \tab The resulting graphs are used as the basis for event extraction. Predicate nodes with PropBank senses are treated as candidate event mentions, and their arguments are passed for ontology role assignment and event typing (shown in~\ref{sec:method:roles} and~\ref{sec:method:lexical}). Each candidate event is scored for confidence at the sentence level, meaning that the confidence in a given scored event is maintained at the span of local evidence (the sentence) before temporal ordering. \\
\paragraph{Role Extraction.}\label{sec:method:roles} \noindent The redaction process outputs a pseudonym map that assigns names to each individual (e.g., \texttt{Victim\_$\texttt{1}$}, \texttt{Suspect\_Unknown}, \texttt{Officer}) and for each sentence in the AMR it gives a variable associated with each name. These are used to ground the predicate--argument roles in the AMR to specific participants that are relevant to the policing domain. \\
\tab For each case we specified a function to map each pseudonym to a unique entity identifier where each case--linked entity is assigned a semantic type (e.g., \cls{Person} or \cls{Vehicle}), the role associated with that mention and sentence--indexed evidence showing where it appears in the narrative. \\
\paragraph{Unknown--actor separation.} \noindent When the pseudonym map shows different unknown suspects in the same case \(c\) they are stored as separate individuals and not merged: 
\[
    \begin{aligned}
        & \PlaysRole(x, \textit{suspect\_unknown}, c) \\
        & \quad \wedge\ \PlaysRole(y, \textit{suspect\_unknown}, c) \\
        & \quad \wedge\ x \neq y \\
        & \qquad \Rightarrow\ \DifferentFrom(x, y).
    \end{aligned}
\]

\subsection{Ontology Construction}\label{sec:method:owl}

\begin{figure}[h]
    \centering
    \includegraphics[width=\linewidth]{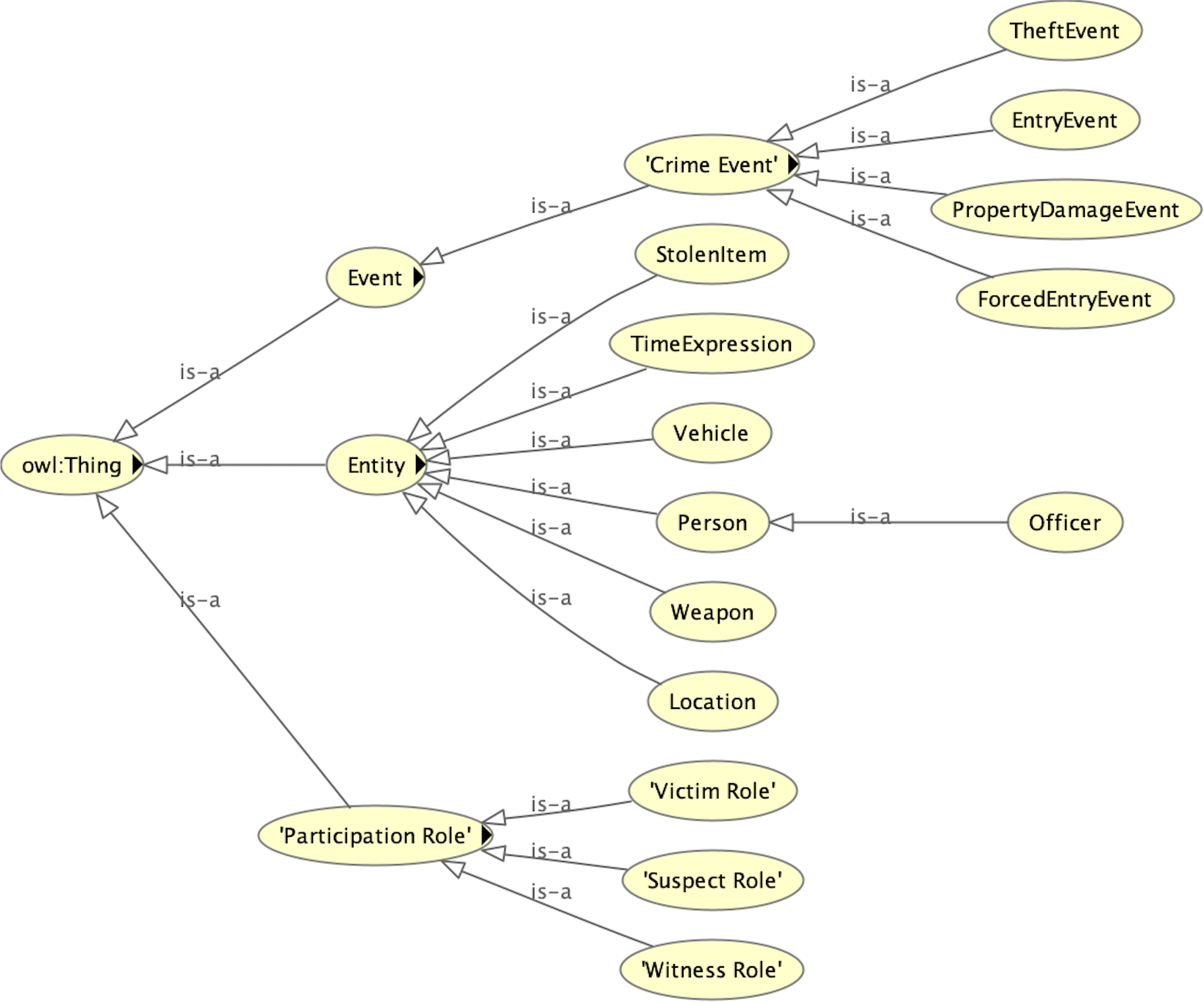}
    \caption{Partial class--level view of the ontology, showing event, entity, and role classes.}
    \label{fig:owl_schema_view}
\end{figure}

\noindent We construct an OWL $2$ ontology in Prot\'eg\'e\footnote{Prot{\'e}g{\'e} is an open--source OWL ontology editor: \texttt{\url{https://protege.stanford.edu}}} to represent case--level knowledge where facts are structured through events and entities. Our ontology uses a relatively compact schema but many case--specific factual assertions. There is a TBox (class schema), an RBox (property schema), and an ABox (case--specific facts). The ontology is dominated by ABox assertions relative to the TBox and RBox. Figure~\ref{fig:owl_schema_view} shows a partial class hierarchy of the ontology which is organized around three main branches: \cls{Event}, \cls{Entity} and \cls{Participation Role} and all descend from \cls{owl}:\cls{Thing}. The \textit{``is--a''} arrows show subclass relationships from general to more specific concepts. For example, \cls{Crime Event} is a subclass of \cls{Event} but \cls{TheftEvent}, \cls{EntryEvent}, \cls{PropertyDamageEvent} and \cls{ForcedEntryEvent} are subclasses of \cls{Crime Event}. \cls{Person}, \cls{Vehicle}, \cls{Weapon} and \cls{Location} are subclasses of \cls{Entity}. \cls{Suspect Role}, \cls{Witness Role} and \cls{Victim Role} are subclasses of \cls{Participation Role}. This class hierarchy helps the system reason at different levels of abstraction, for example by recognizing that an extracted theft event should also satisfy the more general constraints defined for crime events. Ontology statistics are reported in Table~\ref{tab:owl_derived_indicators_app} in Appendix~\ref{app:ontology_metrics}. \\

\paragraph{Ontology structure.} \noindent At the ontology level, the schema includes a set of classes and properties. Most of the properties used at the event scope are related to \cls{Event} participation. Each \cls{Participation} is a relationship between an entity and an event. The property \role{inEvent} captures participants that are part of a given event. Each role in a predicate gives a unique role property (e.g., \role{hasAgent} or \role{hasPatient}) that links the participation to an entity. The property \role{supportedBy} links extracted assertions back to sentences and \role{inCase} associates each event with its report case. \\
\paragraph{Ontology mapping from AMR.} \noindent AMR predicate--argument structures are converted into ontology assertions with a small set of recurring mapping patterns. For example \textit{steal-$01$} \texttt{:ARG0} \texttt{x} \texttt{:ARG1} \texttt{y} is mapped to an event instance \(e\) such that
\[
    \begin{aligned}
        & \cls{TheftEvent}(e) \wedge \exists p\,(\cls{Participation}(p) \wedge \role{inEvent}(p, e) \\
        & \quad \wedge\ \role{hasAgent}(p, x) \wedge \role{hasPatient}(p, y)).
    \end{aligned}
\]
\tab We keep an inspectable intermediate representation of the meaning graphs that includes case--local pseudonyms for entities and concepts, role assignments and the verb together with its numbered sense label for audit of each extracted role and event to verify the corresponding AMR node and sentence text. \\ 
\tab The final lexical typing depends on predicate sense, argument structure and semantic checks over participant types, which is described in Section~\ref{sec:method:lexical}. \\ \\
\noindent \textbf{Description logic (DL).} We apply a small set of DL constraints that capture structural invariants of the ontology. Here $R$ is an object property and $C$ a class; $\sqsubseteq$ means subsumption, $\exists R.C$ an existential restriction, $\forall R.C$ a universal restriction, $\sqcap$ concept intersection and $\top$ the universal class. These are written to capture missing entities, events, links or wrong typed roles. \\
\begin{enumerate}[noitemsep]
    \item \textbf{Participation constraints.} Every participation instance must be linked to an event, and every event must be associated with a case:
    \begin{align*}
        \cls{Participation} & \sqsubseteq \exists\,\role{inEvent}.\cls{Event} \\
        \cls{Event} &\sqsubseteq \exists\,\role{inCase}.\cls{Case}
    \end{align*}

    \item \textbf{Role constraints.} Role assertions are restricted so that only appropriately typed individuals can fill each role:
    \begin{align*}
        \exists\,\role{hasAgent}.\top & \sqsubseteq \cls{Participation} 
    \end{align*}

    \item \textbf{Event constraints.} Some event types have additional requirements on their participants. For example, theft events are modeled as a subclass of \cls{CrimeEvent}, must include a stolen item, and require that the stolen item be an instance of \cls{Item}:
    \begin{align*}
        \cls{TheftEvent} & \sqsubseteq \cls{CrimeEvent} \sqcap \exists\,\role{hasStolenItem}.\cls{Item} \\
        \cls{TheftEvent} & \sqsubseteq \forall\,\role{hasStolenItem}.\cls{Item}
    \end{align*}

    \item \textbf{Domain and range constraints.} The axioms to restrict role properties:
    \[
        \exists\,\role{hasVictim}.\top \sqsubseteq \cls{Event}
    \]
\end{enumerate}

\noindent \tab These are evaluated using the HermiT reasoner~\cite{HermiT}. The axioms check for missing participants, incorrectly typed role fillers and contradictory role assignments.

\subsection{Lexical Rules}\label{sec:method:lexical}

\noindent We represent the meaning of a sentence with predicate--argument structure where the main predicate is the event trigger and the arguments give the initial participant structure. For example the sentence \textit{``The suspect stole a wallet''} gives \(\cls{TheftEvent}(e) \wedge \role{hasAgent}(e, x) \wedge \role{hasPatient}(e, y) \wedge \cls{Suspect}(x) \wedge \cls{Item}(y)\). So \texttt{:ARG0} is the actor and \texttt{:ARG1} is the affected entity. \\
\tab We use three lexical resources for predicate and role normalization. PropBank gives verb senses and their argument roles, which are represented in the AMR graph. SemLink maps each PropBank sense to a VerbNet class which proposes broad candidate event types from the verb or verb family. WordNet gives synsets and a hypernym hierarchy for semantic typing of extracted arguments, such as distinguishing structures (e.g., vehicle) from their parts (e.g., car window). \\
\tab A lemma is the basic dictionary form of a word. For example, \textit{stole}, \textit{steals} and \textit{stealing} all reduce to the lemma \textit{steal}. This begins with PropBank predicate senses, which are mapped through SemLink to VerbNet classes and then to WordNet synsets. Event typing is performed in \textit{two} stages: the predicate--family stage, which generates a coarse set of candidate event types (e.g., \textit{theft}, \textit{entry}, \textit{break}) for a given predicate or predicate family; and the argument--sensitive, refined stage, where the candidate events are filtered, given information about the typed participants and objects. In cases where a sense has no sense mapping, a lemma--level lexical retrieval is performed and this fallback is recorded in the extracted event. \\
\paragraph{WordNet--supported typing.} \noindent We use WordNet to support \textit{``is--a''} checks between arguments and lexical entries where \textit{is--a}$(x,y)$ means that $x$ is a more specific type of $y$ in the hypernym hierarchy~\cite{Schubert2002WorldKnowledge}. For example if a \textit{sedan} is linked through WordNet to \textit{car} and \textit{car} to \textit{vehicle} then \textit{is--a}$(\textit{sedan}, \textit{vehicle})$ gives evidence that the argument belongs to a vehicle class. In \textit{``he broke the window''} we link the object \texttt{window} to the noun synset \texttt{window.n.01} and follow the path and its hypernyms (\texttt{window} $\rightarrow$ \texttt{window.n.01} $\rightarrow$ \texttt{structure\_part}) and determine that window is a structural part. With the break--like predicate sense on the verb get, this supports a \cls{ForcedEntryEvent} or \cls{PropertyDamageEvent}. \\
\paragraph{Argument--sensitive disambiguation.} \noindent PropBank predicate senses alone are not sufficient to give a stable policing event type. So, predicate typing is resolved from the predicate family and the typed argument structure. For example, theft interpretations are strengthened when the affected object belongs to a property--related class and damage interpretations are strengthened when the object is typed as a structure part or vehicle part. This avoids collapsing semantically different cases under the same predicate. \\
\paragraph{Confidence score.} \noindent For each event mention \(e\), we give a score \(c(e) \in [0, 1]\) that says to what extent the assigned event type is supported by the given evidence. This is not a learned probability but an \textit{exploratory heuristic} to summarize the amount of evidence behind an event typing decision. In~\ref{sec:eval:results} the confidence score is treated as a reliability--oriented summary. We report the proportion of events whose confidence goes beyond the given threshold as an indicator of strongly typed extractions. The score is computed from event categories, lexical grounding quality, structural rule support and penalties for ambiguity. \\
\tab We first compute a raw score
\[
    c_{\mathrm{raw}}(e) = b(e) + g_{\mathrm{lex}}(e) + g_{\mathrm{struct}}(e) - p(e),
\]
\noindent where the base term \(b(e)\) comes from the initial policing bucket of the event. Events in the \texttt{incident\_core} group start at \(0.55\), \texttt{police\_action} start at \(0.50\), \texttt{context\_admin} events start at \(0.12\) and uncertain events start at \(0.30\). These initial values give a starting point and judgment about how central the event is to the report. The main incident events start higher and some uncertain events begin lower. \\
\tab The lexical term \(g_{\mathrm{lex}}(e)\) is how strongly the predicate is grounded in the lexical resources. When the event is supported through the full PropBank \(\rightarrow\) VerbNet \(\rightarrow\) WordNet path we add \(+0.25\). When that full path is unavailable and the system falls back to a lemma it adds only \(+0.10\), so stronger lexical mapping gives a larger increase. \\
\tab The term \(g_{\mathrm{struct}}(e)\) measures how well the event matches an expected event pattern. If the predicate has a match for a target event type such as theft, entry or damage then the score increases by \(+0.25\) and object evidence can add \(+0.15\). For example, if the typing rule contains \texttt{obj\_property}  and the object lemmas include property--related words such as \textit{wallet} the score is increased. Also, if the rule contains \texttt{obj\_structure\_or\_vehicle\_part} and the object includes words such as \textit{door}, \textit{window} or other structure or vehicle terms, the score also increases. \\
\tab The penalty term \(p(e)\) reduces the score when the evidence is more ambiguous or uncertain. We subtract a small amount based on the number of WordNet synsets and VerbNet senses associated with the predicate so more lexical ambiguity lowers the confidence. We also subtract \(0.35\) for negation in core incident events or \(0.10\) for negation outside that group, and subtract \(0.12\) for uncertain language such as \textit{appears}, \textit{possibly} or \textit{likely}. \\
\tab When a rule is available, the final score is:
\[
    c(e)=\alpha\,r(\rho_e) + (1 - \alpha)\,\mathrm{bound}_{[0, 1]}\!\big(c_{\mathrm{raw}}(e)\big).
\]
\tab Otherwise:
\[
    c(e)=\mathrm{bound}_{[0, 1]}(c_{\mathrm{raw}}(e)).
\]
\tab Here \(\rho_e\) is the triggered typing rule, \(r(\rho_e)\) is the rule prior and \(\alpha = 0.7\). This means the final score is computed as \(70\%\) rule prior and \(30\%\) evidence--based score. This was done so that when a specific rule strongly supports the typing, that prior has the main influence but it still allows the lexical and structural evidence to adjust the final value. Also, 
\[
    \mathrm{bound}_{[0, 1]}(x)=\max(0,\min(1,x)),
\]
so the final output is between \(0\) and \(1\). \\
\tab Table~\ref{tab:event_conf_examples} shows example extracted events with different confidence scores where higher values of \(c(e)\) mean better supported extractions. For example \textit{kick-$01$} typed as \cls{ForcedEntryEvent}. This event starts at \(0.55\) because it is in the \texttt{incident\_core} bucket. If it is grounded through the full semantic path it gets \(+0.25\), bringing it to \(0.80\). Since \textit{kick} matches a damage--related anchor, it gets another \(+0.25\) which brings it to \(1.05\). Because the object is something like \textit{door} then the rule \texttt{damage\_anchor+obj\_structure\_or\_vehicle\_part} adds \(+0.15\), bringing it to \(1.20\). Then the score is capped at an object--supported rule at \(0.98\) which blends with the rule prior \(0.85\) as \(0.7(0.85)+0.3(0.98)=0.889\) and then adds \(+0.03\) because the rule is highly specific and this gives \(0.919\). \\
\tab Another example is \textit{take-$01$} with \(0.850\) score for \cls{TheftEvent} where the narrative supports theft with stolen items. An event like \textit{leave-$15$} is typed by a \texttt{narrative\_action} that starts lower because it belongs to \texttt{uncertain} bucket (at \(0.30\)) and \textit{discover-$01$} has a score of \(0.522\) because the narrative evidence is less specific.
\begin{table}[h]
    \centering
    \caption{Example extracted events across different scores.}
    \label{tab:event_conf_examples}
    \scriptsize
    \renewcommand{\arraystretch}{1.05}
    \resizebox{\columnwidth}{!}{
    \begin{tabular}{l l r l l}
        \toprule
        \textbf{Predicate} & \textbf{Event class} & \textbf{Conf.} & \textbf{Typing rule} & \textbf{Evidence} \\
        \midrule
        \textit{kick-$01$}      & \cls{ForcedEntryEvent}     & 0.919 & \texttt{damage+obj\_structure}      & door + kicked + structure \\
        \textit{enter-$01$}     & \cls{EntryEvent}           & 0.920 & \texttt{entry}                      & entry point explicitly stated \\
        \textit{leave-$15$}     & \cls{LeaveObjectEvent}     & 0.488 & \texttt{narrative\_action}          & limited structure \\
        \textit{take-$01$}      & \cls{TheftEvent}           & 0.850 & \texttt{theft}                      & stolen property listed \\
        \textit{turn-over-$12$} & \cls{TransferCustodyEvent} & 0.762 & \texttt{owl\_police\_action\_lemma} & lemma$\rightarrow$WN fallback \\
        \textit{discover-$01$}  & \cls{DiscoveryEvent}       & 0.522 & \texttt{narrative\_action}          & weaker narrative support \\
        \bottomrule
    \end{tabular}
    }
\end{table}

\subsection{Temporal Reasoning}\label{sec:method:temporal}

\noindent We construct case temporal graphs for temporal relationships where the nodes in the graphs are individual events, and the edges are precedence relations supported by evidence. We exploit temporal cues derived from the narrative to construct cue--based edges and complement these with domain--specific rules for precedence.

\subsubsection{Timeline Edges.} 

\noindent We aim to establish an explicit timeline of events by using temporal cues such as \textit{then}, \textit{after} and \textit{before} to order closely related event mentions. 
\begin{table}[h]
    \centering
    \caption{Precedence edges added to the temporal graph.}
    \label{tab:temporal_edge_examples}
    \scriptsize
    \renewcommand{\arraystretch}{1.05}
    \resizebox{\columnwidth}{!}{
    \begin{tabular}{lllll}
        \toprule
        \textbf{Source event} & \textbf{Target event} & \textbf{Edge} & \textbf{Support} & \textbf{Evidence} \\
        \midrule
        \cls{ForcedEntryEvent} & \cls{TheftEvent} & \textit{Precedes} & Axiom & forced entry typically precedes theft \\
        \cls{CallEvent} & \cls{ReportTakenEvent} & \textit{Precedes} & Cue / local order & same local narrative sequence \\
        \cls{ReturnEvent} & \cls{DiscoveryEvent} & \textit{Precedes} & Cue / local order & ``upon returning'' \\
        \cls{ArrestEvent} & \cls{BookingEvent} & \textit{Precedes} & Axiom + local order & arrest followed by booking \\
        \bottomrule
    \end{tabular}
    }
\end{table}

\noindent \textbf{Local cue rules.} For adjacent sentences if sentence \(s_{i + 1}\) begins with or has cues such as \textit{then}, \textit{after} or \textit{before}, the system links the event mention selected from sentence \(s_i\) to the event mention in sentence \(s_{i + 1}\) with a \textit{Precedes} edge:
\[
    \textit{Then}(s_{i + 1}) \wedge \textit{Ev}(e_i, s_i) \wedge \textit{Ev}(e_{i + 1}, s_{i + 1}) \to \textit{Precedes}(e_i, e_{i + 1}).
\]
\tab For example \textit{``Suspect (S) entered the home. Then Victim (V) discovered the damage,''} the cue \textit{then} places the discovery event after the entry event. Within a single sentence, cues such as \textit{before} and \textit{after} are used in the same way. And in this example \textit{``the suspect broke the window before entering the home,''} the breaking event is ordered before the entering event. \\ \\
\noindent \textbf{Domain axioms.} We also apply a small set of local domain precedence axioms over event classes (shown in Table~\ref{tab:temporal_edge_examples}). For example if a narrative states that someone broke into a house and later property was taken the system adds an edge placing the forced--entry event before theft event so these axioms add candidate \textit{Precedes} edges:
\[
    \textit{ForcedEntryEvent}(e_f, c)\wedge \textit{TheftEvent}(e_t, c)
    \to \textit{Precedes}(e_f, e_t).
\]

\subsubsection{Temporal Graph Construction.} 

\noindent By constructing typed event nodes, participants and precedence edges, we build temporal graphs that correspond to redacted narratives (see~\ref{sec:eval:results}). The resulting graphs have links to the sentences for verification of which events were extracted from the corresponding AMR and how local cues and domain axioms contributed to event temporal ordering.

%% file: evaluation.tex
\section{Evaluation}\label{eval}

\subsection{Experimental Setup}

\noindent This evaluation measures how well the pipeline structures its output in events, frames and temporal edges. We include a short review to determine the practical usefulness of the symbolic approach. \\
\noindent \tab To address RQ${2}$, we conducted a short questionnaire with $5$ redacted narratives, one from each offense category: \texttt{Burglary}, \texttt{Larceny}, \texttt{Motor Vehicle Theft}, \texttt{Stolen Property} and \texttt{Robbery}, and we asked $6$ reviewers to answer the same $9$ questions for each case. The full questionnaire is provided in Table~\ref{tab:questionnaire} in Appendix~\ref{app:questionnaire}.

\subsection{Metrics}

\noindent We use two complementary forms of evaluation: \\
\paragraph{Corpus--level.} \noindent We report corpus--level results from the symbolic pipeline outputs, including role coverage, event typing, semantic grounding coverage, participant counts, frame slot filling, and temporal edge coverage. \\
\tab The confidence score was measured as:
\[
    \text{HighConf}(\tau) = \frac{|\{e \in \mathcal{E} : c(e) \ge \tau\}|}{|\mathcal{E}|},
\]
where $\mathcal{E}$ is the set of extracted events, $c(e)$ is the confidence score assigned to event $e$ and $\tau$ is a threshold (in our case, $\tau = 0.80$). \\
\tab PB$\rightarrow$VN$\rightarrow$WN coverage was computed as the percentage of extracted events whose semantic mapping followed the full path as well as the lemma$\rightarrow$WN fallback. \\
\tab Participants per case were summarized using the number of unique extracted participants in each case, and we report the median and maximum over all cases. \\
\tab Frame slot filling was measured as the percentage of relevant frames in which a given slot was non--empty. \\
\tab Temporal support was reported in terms of the percentage of cases where at least one temporal edge was extracted with the average number and proportion of cue vs.\ axiom--based edges. 
\begin{table}[h]
    \centering
    \caption{Corpus--level extraction and ordering results.}
    \label{tab:narrative_value_grid}
    \scriptsize
    \setlength{\tabcolsep}{4pt}
    \renewcommand{\arraystretch}{1.03}
    \begin{tabular*}{0.85\columnwidth}{@{\extracolsep{\fill}} l l @{}}
        \toprule
        \textbf{Metric} & \textbf{Value} \\
        \midrule

        \multicolumn{2}{l}{\textbf{Event}} \\
        \hspace{0.6em} Total events & $6686$ \\
        \hspace{0.6em} Total cases & $450$ \\
        \hspace{0.6em} ARG$0$ present & $29.8\%$ \\
        \hspace{0.6em} ARG$1$ present & $78.8\%$ \\
        \hspace{0.6em} Both ARG$0$ and ARG$1$ & $23.4\%$ \\
        \hspace{0.6em} Participants/case (med./max.) & $4$ / $25$ \\
        \hspace{0.6em} Typed with conf.\ $\ge 0.80$ & $54.1\%$ \\
        \hspace{0.6em} PB$ \rightarrow $VN$ \rightarrow $WN & $93.7\%$ \\
        \hspace{0.6em} Lemma$\rightarrow$WN fallback & $6.3\%$ \\

        \midrule
        \multicolumn{2}{l}{\textbf{Frame}} \\
        \hspace{0.6em} Total frames & $6652$ \\
        \hspace{0.6em} Entry point filled & $41.8\%$ \\
        \hspace{0.6em} Entry method filled & $33.4\%$ \\
        \hspace{0.6em} Entry structure filled & $15.5\%$ \\
        \hspace{0.6em} Entry tool filled & $2.6\%$ \\
        \hspace{0.6em} Stolen items filled & $22.8\%$ \\
        \hspace{0.6em} Value mentions filled & $14.8\%$ \\

        \midrule
        \multicolumn{2}{l}{\textbf{Temporal}} \\
        \hspace{0.6em} Total ordering edges & $102$ \\
        \hspace{0.6em} Avg.\ edges/case & $0.23$ \\
        \hspace{0.6em} Cue edges & $21.6\%$ \\
        \hspace{0.6em} Axiom edges & $78.4\%$ \\
        \bottomrule
    \end{tabular*}
\end{table}

\paragraph{Human review.} \noindent The answers from the reviewers were summarized using majority agreement, confidence and ambiguity measures. For questions with a single--choice response, the most frequently selected answer was taken as the summary human label and ties were labeled as \textit{Not clear}. For multiple--choice questions, an option was included only if it received more than a strict majority of reviewer votes, using the threshold $t = \left\lfloor \frac{n}{2} \right\rfloor + 1$. With $n = 6$, this required at least $4$ votes. \\
\tab Human--system agreement was computed only on cases where the human majority was clear. Here, \(n_{\text{match}}\) is the number of matches between the system and the human reference and \(n_{\text{human-clear}}\) is the number of cases for which the human reference was clear:
\[
    \text{Agreement} =
    \frac{n_{\text{match}}}{n_{\text{human-clear}}}\times 100.
\]
\tab We also report majority support as the percentage of response outcomes with a clear majority label, and \textit{Human Not clear} as the percentage of outcomes for which no clear human majority was reached. \\
\tab For the human evaluation, we compared the system’s response against the human majority vote for every question. The \textit{precision} shows how often the system was correct when it predicted that a detail was present, while \textit{recall} shows how often the system found a detail when the human reviewers agreed it was present, and \textit{F$1$} combines precision and recall. These scores were computed only on cases with a clear human majority. For Yes/No questions, we treated the positive case as the presence of the detail.

\subsection{Results}\label{sec:eval:results}

\noindent Table~\ref{tab:narrative_value_grid} shows the corpus--level results over $450$ reports. The symbolic framework produced $6{,}686$ events and $6{,}652$ frames. \texttt{ARG1} coverage was much higher than \texttt{ARG0} ($78.8\%$ vs.\ $29.8\%$) since affected objects are often stated more explicitly than actor identity. The proportion of high--confidence ($\ge$ $0.80$) events was $54.1\%$, and for most typed events, the semantic path used to link them was complete ($93.7\%$). In terms of frame analysis, important details about events, such as the entry point ($41.8\%$), entry method ($33.4\%$), and stolen items ($22.8\%$) were recovered. The most temporal ordering edges were introduced by domain axioms ($78.4\%$). 
\begin{figure}[h]
    \centering
    \includegraphics[width=\columnwidth,height=0.30\textheight,keepaspectratio]{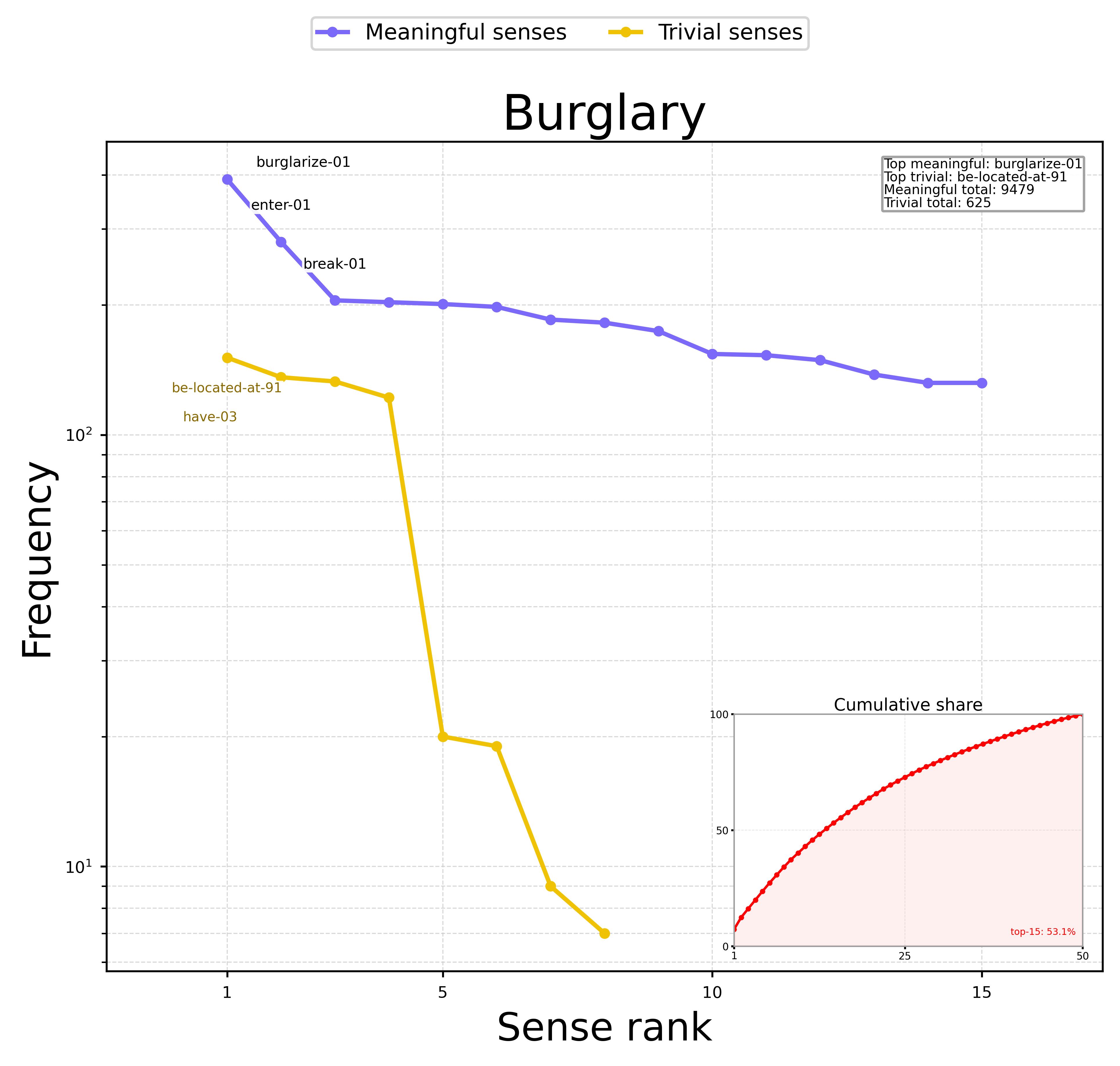}
    \caption{Ranked frequency distribution of extracted AMR predicate senses for \texttt{Burglary}. Purple curves show meaningful senses, yellow curves show trivial senses, and the annotation box summarizes top senses.}
    \label{fig:sense_frequency_panels}
\end{figure}

\noindent \tab To better understand those results Figure~\ref{fig:sense_frequency_panels} shows the ranked frequency distributions of extracted AMR predicate senses for \texttt{Burglary}. The meaningful sense curve shows the predicates with the event content of the narratives and the trivial sense curve shows predicates that are less useful for offense interpretation. In \texttt{Burglary} the most frequent meaningful predicates are \textit{burglarize-$01$}, \textit{enter-$01$} and \textit{break-$01$}. Trivial predicates include forms such as \textit{be-located-at-$91$} and \textit{have-$03$} which are common in AMR graphs and contribute mainly to graph structure. \\
\noindent \tab The burglary sense distribution shows that these narratives have common patterns but also diverse wording. Some predicates appear frequently which give common burglary concepts such as breaking, property loss or damage, however the long tail of lower--frequency senses appear less which shows that these incidents can be described in many different ways. This is important for scaling up to other types of crimes because a keyword--based approach could miss many of these variations and symbolic approach helps by grouping different words under the same event classes and ontology concepts. The plots for \texttt{Larceny}, \texttt{Motor Vehicle Theft}, \texttt{Stolen Property} and \texttt{Robbery} are provided in Figure~\ref{fig:appendix_sense_panels} in Appendix~\ref{app:sense_frequency_four}. 
\begin{figure*}[h]
    \centering
    \includegraphics[width=\linewidth,height=0.32\textheight,keepaspectratio]{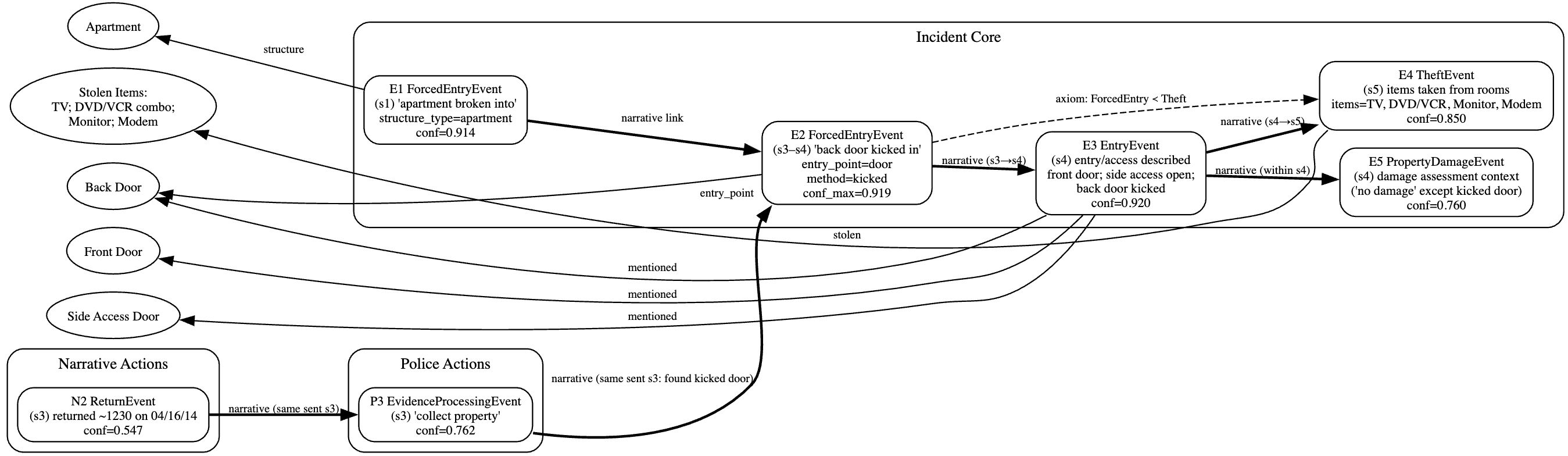}
    \caption{Case--level symbolic graph from a redacted narrative. Solid edges show narrative sequencing and dashed edges show domain axioms.}
    \label{fig:case_graph}
\end{figure*}
\begin{table}[h]
    \centering
    \caption{Human agreement in the five--case review.}
    \label{tab:human_review_agreement}
    \scriptsize
    \setlength{\tabcolsep}{3pt}
    \renewcommand{\arraystretch}{1.02}
    \begin{tabular}{lccc}
        \toprule
        \textbf{Question} & \textbf{Human--clear cases} & \textbf{Agreement} & \textbf{Human Not clear} \\
        \midrule
        Q$1$ Incident initiation & $3$ & $100.0\%$ & $40.0\%$ \\
        Q$2$ Vehicle involvement & $5$ & $80.0\%$ & $0.0\%$ \\
        Q$3$ Forced entry & $5$ & $40.0\%$ & $0.0\%$ \\
        Q$4$ Entry point / damage & $0$ & -- & $100.0\%$ \\
        Q$5$ Theft stated & $5$ & $80.0\%$ & $0.0\%$ \\
        Q$6$ Stolen items named & $5$ & $100.0\%$ & $0.0\%$ \\
        Q$7$ Time cue present & $5$ & $100.0\%$ & $0.0\%$ \\
        Q$8$ Participant roles & $2$ & $100.0\%$ & $60.0\%$ \\
        Q$9$ Overall confidence & $5$ & $80.0\%$ & $0.0\%$ \\
        \bottomrule
    \end{tabular}
\end{table}
\noindent \\ \tab Agreement between the system and human majority for cases where reviewers reached a clear majority in Table~\ref{tab:human_review_agreement}) was $100\%$ for incident initiation (Q$1$), stolen items (Q$6$), time cues (Q$7$) and participant roles (Q$8$). Agreement was $80.0\%$ for vehicle involvement (Q$2$), theft stated (Q$5$), answerability (Q$9$) and $40.0\%$ for forced entry (Q$3$). Ambiguity was highest for entry point (Q$4$) where $100.0\%$ of the cases were \textit{Not clear} (see Figure~\ref{fig:human_review_ambiguity_app} in Appendix~\ref{app:human_review_ambiguity}). \\
\noindent \tab Precision, recall and F$1$ against human--majority were $75.0/100.0/85.7$ for vehicle involvement, $50.0/50.0/50.0$ for forced entry, $100.0/80.0/88.9$ for theft stated, $100.0/100.0/100.0$ for stolen items and time cues and $80.0/80.0/80.0$ for participant roles. Forced entry was challenging where in some cases the pipeline identified forced entry but reviewers marked that evidence as uncertain (Table~\ref{tab:human_review_prf}). 
\begin{table}[h]
    \centering
    \caption{Precision, recall, and F1 against human--majority votes.}
    \label{tab:human_review_prf}
    \scriptsize
    \setlength{\tabcolsep}{4pt}
    \renewcommand{\arraystretch}{1.03}
    \begin{tabular}{lcccc}
        \toprule
        \textbf{Metric} & \textbf{Cases} & \textbf{Precision} & \textbf{Recall} & \textbf{F1} \\
        \midrule
        Q2 Vehicle involved & 5 & 75.0\% & 100.0\% & 85.7\% \\
        Q3 Forced entry & 4 & 50.0\% & 50.0\% & 50.0\% \\
        Q5 Theft stated & 5 & 100.0\% & 80.0\% & 88.9\% \\
        Q6 Stolen items & 5 & 100.0\% & 100.0\% & 100.0\% \\
        Q7 Specific time cue & 5 & 100.0\% & 100.0\% & 100.0\% \\
        Q8 Participant roles & 2 & 80.0\% & 80.0\% & 80.0\% \\
        \bottomrule
    \end{tabular}
\end{table}

\noindent \tab Figure~\ref{fig:case_graph} shows a partial view of a case temporal graph. The output is mapped to police actions, narrative actions and main incident events capturing the language of building entry point, the specific items that were stolen and the local ordering of events. The full graph is provided in Figure~\ref{fig:full_case_graph_appendix} in Appendix~\ref{app:full_case_graph}. \\
\noindent \tab Our findings support both research questions. For \textbf{RQ$\textbf{1}$}, our symbolic pipeline is able to extract evidence--linked facts from cases and relate them to the ontology for reasoning. In many cases, the pipeline is able to recover temporal facts beyond, e.g., for theft, it is able to recover facts such as what was stolen and how the events of a case are sequenced. Reviews of redacted narratives for \textbf{RQ$\textbf{2}$} show that many questions about entities, events, and roles can still be answered from redacted narratives.

%% file: discussion.tex
\section{Discussion}\label{limits}

\noindent Police narratives can be different in completeness, wording and style, so redaction, OCR or the use of abbreviations such as \texttt{V}, \texttt{S} or \texttt{W} can make extraction less reliable. Moreover, the use of anonymized references might lead to confusion in the interpretation of descriptive phrases that are meant to refer to unknown persons. \\
\tab The environment used in this work does not have a gold standard corpus for events, participants, and temporal ordering, so the evaluation focuses on traceability and not gold--standard benchmark performance. \\
\tab The intended use of this pipeline is to extract entities, events, roles and their ordering. However, it should not be used for making conclusions about investigations, ranking police officers or deciding about individuals without human review. \\
\tab In the future we will focus on improving entity linking, so that repeated mentions across reports are more reliably mapped to the same people, vehicles, or items. We will also compare the symbolic approach with the outputs produced by a large language model.

%% file: conclusion.tex
\section{Conclusion}\label{conclude}

\noindent Police narratives contain important details beyond the information from structured fields. Incident details can be recovered using a symbolic ontology approach, augmented with AMR, PropBank, VerbNet and WordNet. Results show that for the extracted events, $54.1\%$ have a confidence score $\ge$ $0.80$ and $93.7\%$ were mapped through the full semantic hierarchy. Human reviewers had $100\%$ agreement for incident initiation, stolen items and specific time cues. However, temporal ordering was the most challenging for forced entry signal. Overall, these results show that symbolic NLU can show uncertainty through confidence scores as well as keep traceability through lexicon and sentence evidence.

%% file: acknowledgments.tex
\section{Acknowledgments}\label{acknowledgments}

\noindent I thank my faculty advisor Dr. Jansen Orfan, and the Rochester (NY) Police Department Office of Business Intelligence for the data, guidance and support.

%% file: appendix.tex
\section*{Appendix}\label{Appendix}

\section{Supplementary Ontology Statistics}\label{app:ontology_metrics}

\begin{table}[H]
    \centering
    \caption{Derived ontology indicators from Prot\'eg\'e metrics.}
    \label{tab:owl_derived_indicators_app}
    \scriptsize
    \setlength{\tabcolsep}{3pt}
    \renewcommand{\arraystretch}{1.05}
    \begin{tabular}{l p{0.50\linewidth} r}
        \toprule
        \textbf{Symbol} & \textbf{Definition} & \textbf{Value} \\
        \midrule
        $N_{\mathrm{ax}}$ & Total axioms & $192{,}296$ \\
        $N_L$ & Logical axioms & $153{,}028$ \\
        $N_D$ & Declaration axioms & $39{,}249$ \\
        $N_{\mathrm{ann}}$ & Annotation assertions & $19$ \\
        \midrule
        $r_L$ & Logical share: $r_L = \frac{N_L}{N_{\mathrm{ax}}}$ & $79.6\%$ \\
        $r_D$ & Declaration share: $r_D = \frac{N_D}{N_{\mathrm{ax}}}$ & $20.4\%$ \\
        $r_{\mathrm{ann}}$ & Annotation share: $r_{\mathrm{ann}} = \frac{N_{\mathrm{ann}}}{N_{\mathrm{ax}}}$ & $0.01\%$ \\
        \midrule
        $N_{\mathrm{CA}}$ & ClassAssertion count & $39{,}153$ \\
        $N_{\mathrm{OPA}}$ & ObjectPropertyAssertion count & $39{,}785$ \\
        $N_{\mathrm{DPA}}$ & DataPropertyAssertion count & $73{,}964$ \\
        $N_{\mathrm{ABox}}$ & Assertion axioms: $N_{\mathrm{CA}} + N_{\mathrm{OPA}} + N_{\mathrm{DPA}}$ & $152{,}902$ \\
        $\rho$ & Assertion density: $\rho = \frac{N_{\mathrm{ABox}}}{N_{\mathrm{ind}}}$ & $3.91$ \\
        $\bar d_{\mathrm{obj}}$ & Avg.\ obj.\ assertions/individual: $\bar d_{\mathrm{obj}} = \frac{N_{\mathrm{OPA}}}{N_{\mathrm{ind}}}$ & $1.02$ \\
        $\bar d_{\mathrm{data}}$ & Data assertions/individual: $\bar d_{\mathrm{data}} = \frac{N_{\mathrm{DPA}}}{N_{\mathrm{ind}}}$ & $1.89$ \\
        \midrule
        $N_C$ & Number of classes & $71$ \\
        $N_{\mathrm{ind}}$ & Number of individuals & $39{,}153$ \\
        $\kappa$ & Individuals per class: $\kappa=\frac{N_{\mathrm{ind}}}{N_C}$ & $551.5$ \\
        \midrule
        $c^{\mathrm{obj}}_{\mathrm{dom}}$ & Obj.\ domain coverage: $\frac{N^{\mathrm{obj}}_{\mathrm{dom}}}{P_{\mathrm{obj}}}=\frac{20}{20}$ & $100\%$ \\
        $c^{\mathrm{obj}}_{\mathrm{rng}}$ & Obj.\ range coverage: $\frac{N^{\mathrm{obj}}_{\mathrm{rng}}}{P_{\mathrm{obj}}}=\frac{20}{20}$ & $100\%$ \\
        $c^{\mathrm{data}}_{\mathrm{dom}}$ & Data domain coverage: $\frac{N^{\mathrm{data}}_{\mathrm{dom}}}{P_{\mathrm{data}}}=\frac{6}{6}$ & $100\%$ \\
        $c^{\mathrm{data}}_{\mathrm{rng}}$ & Data range coverage: $\frac{N^{\mathrm{data}}_{\mathrm{rng}}}{P_{\mathrm{data}}}=\frac{6}{6}$ & $100\%$ \\
        \bottomrule
    \end{tabular}
\end{table}

\FloatBarrier

\section{Human Review Questionnaire}\label{app:questionnaire}

\begin{table}[H]
    \centering
    \caption{Questionnaire used for human review of redacted narratives.}
    \label{tab:questionnaire}
    \scriptsize
    \setlength{\tabcolsep}{3pt}
    \renewcommand{\arraystretch}{1.09}
    \begin{tabular}{p{0.29\columnwidth} p{0.65\columnwidth}}
        \toprule
        \textbf{Question} & \textbf{Response options} \\
        \midrule
        
        \textbf{Q$\textbf{1.}$} How was the incident first initiated or reported? & $911$ call; $311$ call; Officer dispatched; Walk--in / in--person report; Officer observed it directly; Other; Not clear. \\
        
        \textbf{Q$\textbf{2.}$} Is a vehicle involved in the incident? &
        Primary vehicle incident; Vehicle mentioned only; No vehicle mentioned; Not clear. \\
        
        \textbf{Q$\textbf{3.}$} Does the narrative indicate that someone forced entry (e.g., door/window was kicked in, pried, broken, smashed)? & Yes; No; Not clear. \\
        
        \textbf{Q$\textbf{4.}$} If Q$3$ is `Yes', what was the entry point/damage? & Door; Window; Glass; Vehicle part (e.g., car window); Other. \\
        
        \textbf{Q$\textbf{5.}$} Does the narrative clearly state that property was stolen/taken? & Yes; No; Not clear. \\
        
        \textbf{Q$\textbf{6.}$} What items were stolen, as stated in the narrative? & If none are named, write None named. If unclear, write Not clear. \\
        
        \textbf{Q$\textbf{7.}$} Does the narrative include any time cue for when the incident happened? & Specific time; Rough time only; No time mentioned; Not clear. \\
        
        \textbf{Q$\textbf{8.}$} Which of these are clearly mentioned in the narrative? & Victim (V) / caller / reporter; Officer / reporting officer; Suspect (S) / unknown person; Person with Knowledge (PK); Witness (W); Other. \\
        
        \textbf{Q$\textbf{9.}$} The narrative provided enough evidence to answer: & All of the questions; Most of the questions; Some of the questions; None of the questions. \\
        \bottomrule
    \end{tabular}
\end{table}

\FloatBarrier

\section{Additional sense--frequency plots}\label{app:sense_frequency_four}

\begin{figure}[H]
    \centering
    \includegraphics[width=\columnwidth]{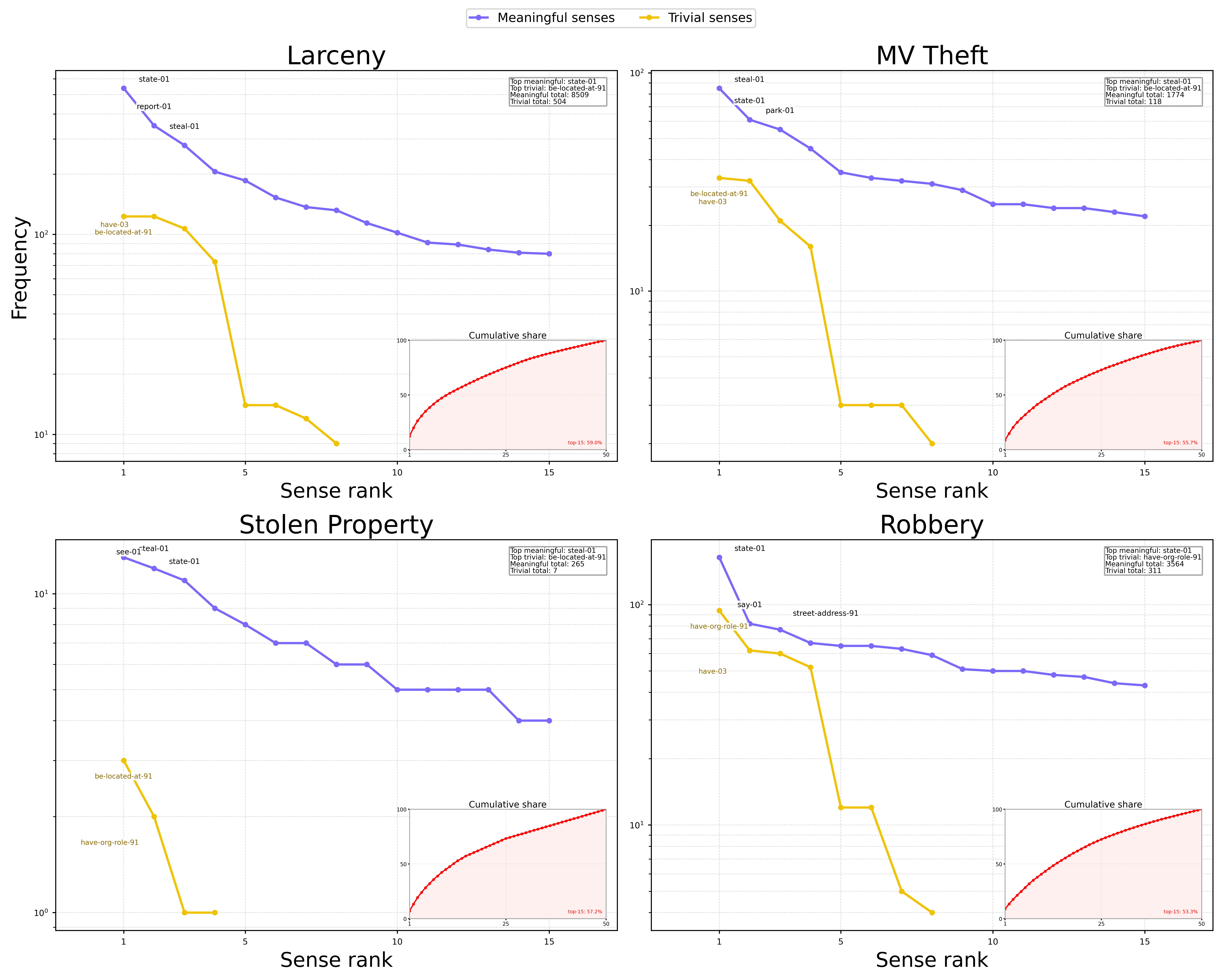}
    \caption{Ranked frequency distributions of extracted AMR predicate senses for \texttt{Larceny}, \texttt{Motor Vehicle Theft}, \texttt{Stolen Property}, and \texttt{Robbery}.}
    \label{fig:appendix_sense_panels}
\end{figure}

\FloatBarrier

\section{Human Review Ambiguity}\label{app:human_review_ambiguity}

\begin{figure}[h]
    \centering
    \begin{tikzpicture}
        \begin{axis}[
                width=\columnwidth,
                height=5.9cm,
                xbar,
                xmin=0, xmax=109,
                bar width=3.5pt,
                xlabel={\%},
                xlabel style={font=\scriptsize, yshift=4.5pt},
                tick label style={font=\scriptsize},
                label style={font=\scriptsize},
                symbolic y coords={
                    Q9 Confidence,
                    Q8 Roles,
                    Q7 Time cue,
                    Q6 Items named,
                    Q5 Theft stated,
                    Q4 Entry point,
                    Q3 Forced entry,
                    Q2 Vehicle,
                    Q1 Initiation
                },
                ytick=data,
                y dir=reverse,
                legend style={
                    at={(0.5,-0.13)},
                    anchor=north,
                    legend columns=2,
                    draw=none,
                    font=\scriptsize
                },
                nodes near coords,
                every node near coord/.append style={font=\tiny},
                nodes near coords align={horizontal},
                enlarge y limits=0.04,
                grid=major,
                grid style={dashed,gray!25},
            ]
            \addplot[fill=myblockcolor!75, draw=myblockcolor] coordinates {
                (76.7,Q1 Initiation)
                (86.7,Q2 Vehicle)
                (86.7,Q3 Forced entry)
                (40.0,Q4 Entry point)
                (96.7,Q5 Theft stated)
                (100.0,Q6 Items named)
                (83.3,Q7 Time cue)
                (30.0,Q8 Roles)
                (76.7,Q9 Confidence)
            };
            \addplot[fill=myblockcolor!25, draw=myblockcolor!70] coordinates {
                (40.0,Q1 Initiation)
                (0.0,Q2 Vehicle)
                (0.0,Q3 Forced entry)
                (100.0,Q4 Entry point)
                (0.0,Q5 Theft stated)
                (0.0,Q6 Items named)
                (0.0,Q7 Time cue)
                (60.0,Q8 Roles)
                (0.0,Q9 Confidence)
            };
            \legend{Majority support, No clear majority}
        \end{axis}
    \end{tikzpicture}
    \caption{Reviewer ambiguity in the five--case review.}
    \label{fig:human_review_ambiguity_app}
\end{figure}
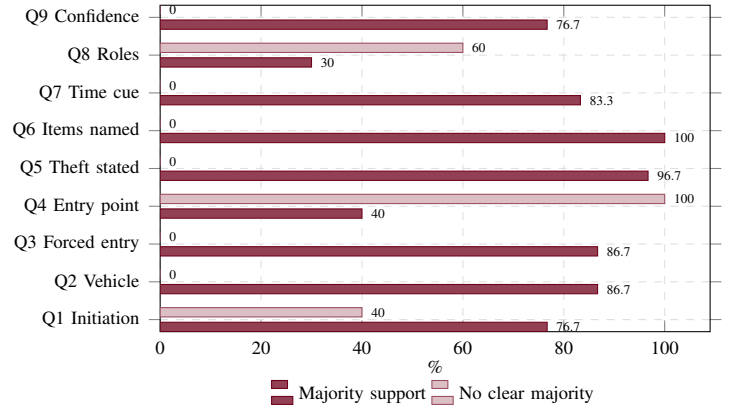

\FloatBarrier

\clearpage
\section{Temporal Case Graph}\label{app:full_case_graph}

\begin{figure}[H]
    \centering
    \includegraphics[
        angle=90,
        origin=c,
        width=\columnwidth,
        height=0.60\textheight,
        keepaspectratio
    ]{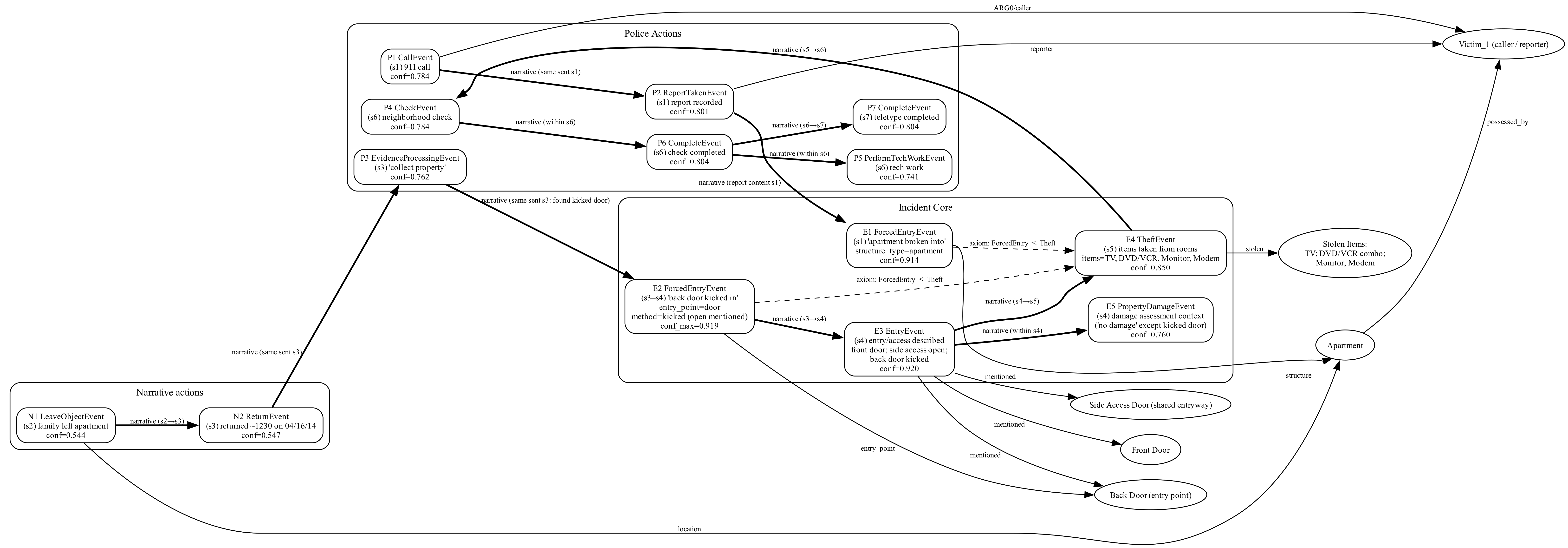}
    \caption{Full temporal case graph from an example redacted report, showing police actions, narrative actions, core incident events, entity links, and temporal relations.}
    \label{fig:full_case_graph_appendix}
\end{figure}

\clearpage